\documentclass[conference]{IEEEtran}

\IEEEoverridecommandlockouts

\usepackage{amsmath}
\usepackage{amsfonts}
\usepackage{amssymb}
\usepackage{graphicx}
\usepackage[colorlinks=true]{hyperref}
\usepackage{pgfplots}
\pgfplotsset{compat=1.17}
\usepackage{dirtytalk}
\usepackage{cite}
\usepackage{microtype}
\usepackage{float}
\usepackage{subcaption}
\usepackage{caption}

\hypersetup{
    linkcolor=red,
    urlcolor=cyan,
    citecolor=blue
}

\begin{document}

\title{
    Cold-Start Active Preference Learning in Socio-Economic Domains
}

\author{
    \IEEEauthorblockN{Mojtaba Fayazbakhsh\IEEEauthorrefmark{1}, Danial Ataee\IEEEauthorrefmark{2}, MohammadAmin Fazli\IEEEauthorrefmark{1}}
    \IEEEauthorblockA{\IEEEauthorrefmark{1}Department of Computer Engineering, Sharif University of Technology}
    \IEEEauthorblockA{\IEEEauthorrefmark{2}Department of Mathematical Sciences, Sharif University of Technology}
}

\maketitle

\begin{abstract}
Active preference learning offers an efficient approach to modeling preferences, but it is hindered by the cold-start problem, which leads to a marked decline in performance when no initial labeled data are available. While cold-start solutions have been proposed for domains such as vision and text, the cold-start problem in active preference learning remains largely unexplored, underscoring the need for practical, effective methods. Drawing inspiration from established practices in social and economic research, the proposed method initiates learning with a self-supervised phase that employs Principal Component Analysis (PCA) to generate initial pseudo-labels. This process produces a \say{warmed-up} model based solely on the data's intrinsic structure, without requiring expert input. The model is then refined through an active learning loop that strategically queries a simulated noisy oracle for labels. Experiments conducted on various socio-economic datasets, including those related to financial credibility, career success rate, and socio-economic status, consistently show that the PCA-driven approach outperforms standard active learning strategies that start without prior information. This work thus provides a computationally efficient and straightforward solution that effectively addresses the cold-start problem.

\end{abstract}

\renewcommand\IEEEkeywordsname{Index Terms}

\section{Introduction}

Comprehensive analysis in social and economic research often requires integrating multiple factors into assessments to capture the complexity of real-world phenomena. For example, to gauge overall societal progress effectively, the UK Government's guidance on appraisal emphasizes the importance of considering a range of indicators beyond mere economic outputs, such as health, education, and environmental quality, \cite{maxwell2011social}. Another instance is the literature on measuring and analyzing poverty, given its multidimensional nature (e.g., \cite{alkire2015multidimensional,d2024multidimensional}). While defining simple, theoretically interpretable evaluation criteria is sometimes infeasible, formal preference models address the challenges of preference elicitation and provide a way for a decision maker to express their assessment in a clear, declarative manner.

Developing a preference model over alternatives has a long history with analogous terminology across diverse disciplines, including machine learning, artificial intelligence, operations research, the social sciences, economics, and decision theory \cite{furnkranz2010preference}. Revealed preference theory, introduced by economist Paul Samuelson in 1938 \cite{samuelson1938pure,samuelson1948consumption}, asserts that consumer preferences can be inferred from their purchasing decisions, assuming rational behavior aimed at maximizing utility given constraints in terms of price and income \cite{cozic2025roles}. This approach emphasizes that actual choices made by consumers better reflect their true preferences than self-reported data, which can be biased or unreliable. 

Multi-criteria decision making (MCDM), a known term in operational research and decision sciences \cite{thakkar2021multi}, focuses on providing disciplined methods to structure complex decision problems and evaluate alternatives concerning multiple conflicting aspects \cite{taherdoost2023multi}, for example, cost and quality \cite{shahsavarani2015bases}. Systematic reviews have shown robust growth for applying MCDM methods \cite{taherdoost2023multi} in finance, transportation, medicine, management, engineering, and sustainable development studies \cite{aruldoss2013survey,pramanik2021comparative}. The process of multi-criteria decision making generally involves two phases: evaluating alternatives under different subcriteria and combining evaluation vectors into a final comparable score for each item \cite{aggarwal2019modelling}. While there are recent efforts to exploit machine learning techniques \cite{alves2023machine,meng2024multi}, the aggregation of subcriteria-wise evaluations (i.e., the core of the preference model) is usually done by each method's proposed rule.

In machine learning, preference learning is considered a task that focuses on the automated acquisition of preference models from observed or revealed preference information. Preference learning involves predicting preferences over a collection of alternatives \cite{aggarwal2019modelling}. These preferences can be represented in various forms, such as utility functions, preference relations, or logical representations. The task is to learn these preferences from empirical data, which may include pairwise comparisons or other types of preference information \cite{chen2024preference,hullermeier2008label}. Similar to revealed preferences and consumer theory, preference learning uses algorithms to analyze past preference outcomes to predict future choices, effectively learning the underlying preferences from observed actions rather than hypothetical scenarios \cite{beshears2008preferences}. Both frameworks operate on the principle that real-world data provides a more reliable foundation for understanding consumer behavior than theoretical constructs alone. Thus, the intersection of revealed preferences and preference learning highlights a shared goal: to derive insights into decision-making by focusing on actual made choices rather than assumptions about utility or comparison mechanisms.

One approach to preference learning is to use utility functions. These functions assign an absolute degree of utility to each alternative, enabling ranking based on their utility scores. Learning utility functions can be framed as a regression problem, depending on whether the utility scale is numerical or ordinal. Another approach involves learning binary preference relations, which predict whether one alternative is preferred over another. This method reduces preference learning to a binary classification problem, where the goal is to learn a preference predicate that compares pairs of alternatives. This approach is more general and can handle partial-order relations, but it may lead to inconsistencies, such as preferential cycles \cite{furnkranz2010preference}.

One of the primary challenges in machine learning is the labeling process. Labeling, especially in socio-economic domains, is often difficult because these problems are complex and multifaceted, and expert knowledge alone may not be sufficient to provide accurate and comprehensive labels for training data. Furthermore, experts may differ in their interpretations of the same problem and disagree on what constitutes the \say{correct} label, introducing subjectivity and inconsistencies into the dataset. This can be especially problematic when the domain involves ethical, cultural, or political considerations that are difficult to quantify. This subjectivity can introduce biases into the data, which might not be immediately apparent but can significantly affect model performance and fairness \cite{mehrabi2019survey}. Experts are often influenced by their personal experiences and societal norms, which can result in biases that go unnoticed during the labeling process \cite{binns2018fairness}. These biases, once embedded in the training data, can lead to biased predictions from the model, especially in socio-economic domains, where the stakes for fairness and equality are high.

Moreover, the resource-intensive nature of labeling large datasets further complicates the process, as it requires expert time and knowledge, which may be limited or not readily available. This situation becomes even more difficult when data is incomplete or noisy, forcing experts to make judgments based on imperfect information, thereby degrading the quality of the labeled data.

The above-mentioned concerns underscore the importance of methods to address expert biases and reduce reliance on large, inconsistent labeled datasets, enabling more scalable, efficient learning in socio-economic contexts. Some researchers have proposed evaluation metrics derived from unsupervised techniques, such as principal component analysis, to avoid labeling challenges \cite{filmer2001estimating,reid2024spatially,tkach2022multidimensional,lamichhane2021benchmarking}. Surprisingly, these methods have shown acceptable performance in practice, leading to widespread adoption in some communities \cite{gwatkin2007socio}. However, this approach has some drawbacks. First of all, it seems there is a lack of theoretical justification for the approach, a point sometimes acknowledged by the proposers themselves \cite{bollen2007socio}. In addition, full elimination of expert supervision could be overly incautious. Depriving preference models of any guidance limits their performance.

To mitigate the aforementioned challenges, one approach is to simplify the labeling task and align it with human cognitive capabilities. Humans often excel at relative comparisons rather than absolute evaluations. By framing labeling tasks as comparative judgments, such as pairwise comparisons of data points, we can leverage this inherent human strength \cite{kadiouglu2022sample,furnkranz2010preference,ammar2011ranking}. This approach aligns with Thurstone's Law of Comparative Judgment, which posits that subjective judgments can be placed through repeated pairwise comparisons \cite{thurstone1927law}. Adapting the labeling process to focus on relative assessments can lead to more consistent and reliable annotations, reducing the impact of individual biases and improving overall data quality. Furthermore, such simplification can decrease the cognitive load on experts, potentially reducing labeling time and associated costs \cite{miller1956magical}.

Beyond simplifying the labeling process, it is crucial that machine learning models be tolerant of the inevitable errors introduced by expert annotators. Even with simplified tasks and careful training, experts are still prone to occasional mistakes due to fatigue, misinterpretation, or inherent ambiguity in the data. Therefore, models should be designed and trained to perform effectively even when supervised by a \say{noisy oracle} – a labeling source that provides generally accurate but imperfect information \cite{angluin1988learning,northcutt2021confident,zhang2018generalized}.

To further address the challenges of cost, time, and biases inherent in the labeling process, integrating human-in-the-loop (HITL) methodologies—particularly active learning—can significantly enhance the efficiency and quality of supervised learning systems \cite{wing2021trustworthy,yang2021human}. Strategic active learning involves human annotators by iteratively selecting the most informative samples for labeling, thereby reducing the number of annotations required while maintaining, or even improving, model performance. This approach not only minimizes the labeling burden but also ensures that human expertise is utilized where it is most impactful, such as in resolving ambiguous or uncertain cases. By incorporating active learning and HITL principles, the labeling process becomes more efficient, cost-effective, and less prone to errors and biases, ultimately leading to more robust and generalizable machine learning models \cite{mosqueira2023human}.

Active learning typically depends on the availability of initial labeled data. A major challenge in this context is the cold-start problem, which arises when a model must select query samples with little or no prior labeled data. Cold-start active learning strategies are designed to provide an effective warm-up and rapidly accumulate sufficient labeled data to initiate the learning process. Recent state-of-the-art methods employ self-supervised techniques to generate informative queries during the cold-start phase \cite{yuan2020cold}. These approaches are particularly advantageous when no initial labeled samples are available and the annotation budget is limited, which is precisely when active learning is most valuable. Proper model initialization has emerged as a best practice in active learning \cite{lang2021best}. Notably, there is currently no original research addressing the cold-start issue in active preference learning. Leveraging widely used data analysis techniques from social studies, this work introduces a method that outperforms baseline approaches in the early stages. To ensure a realistic experimental setting, the models interact with a noisy, probabilistic simulated oracle during the active learning phase.

\section{Previous Works} \label{sec:prev}

The field of preference learning, which seeks solutions to the problem of ordering items based on user preferences, overlaps with the field of learning to rank \cite{chen2019foundations,werner2022review}, which can be traced back to 1940 \cite{kendall1940method}. In recent decades, researchers proposed different techniques, such as extending traditional Support Vector Machines (SVM) to a ranking setting \cite{herbrich2000large,joachims2002optimizing}, leveraging the eigenvectors of certain matrices built from the data by spectral methods \cite{fogel2016spectral,chau2022spectral}, or using boosting algorithms \cite{freund2003efficient,xu2007adarank,chen2016xgboost}.

A major line of effort has come to exploit neural networks. One early contribution was the development of the RankNet algorithm, a pairwise ranking approach based on neural networks \cite{burges2005learning}. RankNet uses the cross-entropy loss function to learn relative rankings between two items, which was pivotal in advancing the field, particularly for applications in recommender systems and search engine ranking. Following them, \cite{burges2006learning} proposed LambdaRank, an extension of gradient-based learning methods tailored for ranking. By directly optimizing ranking metrics such as Normalized Discounted Cumulative Gain (NDCG), LambdaRank advanced the practical application of preference learning algorithms, especially in search engine ranking and information retrieval systems. Other works proposed particular architectures to naturally implement the symmetries present in a preference function \cite{rigutini2011sortnet}. With the rise of graph neural networks, preference learning benefited from looking at preference relations as a directed graph \cite{he2022gnnrank}. Other recent works \cite{zhu2022personalized,zhang2023vae} introduced transfer learning approaches for preference learning, aiming to transfer knowledge learned in one domain to another. This is particularly useful in settings where obtaining labeled preference data is costly or time-consuming.

Research with more realistic assumptions has incorporated handling noisy and probabilistic comparisons. In real-world scenarios, preferences are often uncertain and subject to noise due to imperfections in data collection or ambiguous preferences expressed by humans. As a result, a significant portion of research has focused on extending existing preference models to handle noisy or uncertain pairwise comparisons. These models aim to improve the robustness of the learning process when some of the preference labels are corrupted, missing, or otherwise unreliable. The Bradley-Terry model, which was one of the first models to formalize pairwise preference learning, assumes that the probability of one item being preferred over another is determined by the ratio of their internal score \cite{bradley1952rank}. This basic model and its variations are widely used for probabilistic comparisons \cite{maystre2017just} and are known as a standard assumption for ground-truth human preferences \cite{chen2024preference}. Meanwhile, other modelings for noisy comparisons could be addressed \cite{braverman2008noisy,jamieson2011active,ailon2012active}.

As we mentioned in the previous section, obtaining labeled preference data might be expensive or time-consuming. Active learning is a well-known framework that reduces labeling costs \cite{mosqueira2023human,ijcai2021p634,lang2021best}. The key idea behind active learning in the context of preference learning is to actively select the most informative comparisons to query, thereby reducing the number of required preference judgments while achieving satisfactory accuracy levels \cite{qian2013active,cai2015active,long2010active,donmez2009active}. When dealing with active learning, noisy comparisons are double trouble, since they not only affect the trained model but also the query strategy, which is affected by wrong labels \cite{younesian2021qactor}. A parallel trend has investigated active preference learning from a transductive perspective \cite{maystre2017just,ailon2012active,wang2014active,saha2019active}; they produce a total preference order over a specific set of elements via online queries, but cannot generalize to unseen elements.

However, even active learning algorithms generally rely on initial labeled data for appropriate functioning \cite{ijcai2021p634}. Without prior labeled data, the performance of the active learning process declines because the model struggles to assess the informativeness of samples and select appropriate queries. This problem may be compared with the well-studied issue of \say{cold start} in the recommender systems literature \cite{lika2014facing}, since query selection is essentially a sample recommendation. Interesting benchmarks demonstrate the fact that in the absence of prior labeled data, even naive uniform query selection outperforms famous active query selection strategies when the annotation budget is small \cite{simeoni2021rethinking,pourahmadi2021simple,chandra2021initial,zhu2019addressing,ash2020warmstart}. While using independent and identically distributed (i.i.d) sampling and maintaining a similar distribution between training and test data can be beneficial for some reasons \cite{jadon2021covid}, most of the query strategies are highly biased to outlier samples and specific classes, leading to more generalization error and a vicious cycle caused by poor sample selection for next iterations \cite{chen2023making}. It is common to gather first labeled samples at random. However, going through the \say{inspect \& adapt} loop for query selection is advantageous once sufficient labeled data is available. Gao et al. have studied the optimal turning point for shifting to learning-based sampling methods when facing cold-start problems \cite{gao2020consistency}.

In recent years, cold-start active learning strategies have been proposed to overcome the initial data scarcity and quickly gather sufficient labeled data to bootstrap the learning process. Mentioned works focus on using domain-specific techniques to leverage existing knowledge from similar contexts and enhance initial model performance in an unsupervised manner. For example, pre-trained language models are used to identify surprising samples that should be labeled during the cold start phase, thereby reducing annotation costs while improving text classification performance \cite{yuan2020cold,guo2024deuce}. Zheng et al. \cite{zheng2019biomedical,zheng2020annotation} have proposed a representation-based one-shot active learning framework, composed of a variational autoencoder for feature extraction and a sampling module based on k-means and max-cover algorithms, to perform medical image-related tasks. Another method, addressing natural image classification, has applied contrastive self-supervised learning and hierarchical clustering, achieving up to a 10\% performance increase under low-budget conditions \cite{jin2022cold}. Another method that seeks to maximize probability coverage by using the union of balls of the same radius has shown improvements on image recognition benchmarks \cite{yehuda2022active}.

To the best of our knowledge, there is no original research in the AI community addressing the cold-start issue in preference learning. By the way, there is a similar literature in microeconomics, with the key terms \say{rationalizability} and \say{revealed preferences}. The main idea indicates that each actual choice of a rational decision maker is expected to obey a consistent internal utility function, and we can infer a consumer’s preferences by observing their actual choices between budget-consistent alternatives \cite{samuelson1938note}. The theory emphasizes inducting the preference model from particular relative choices, rather than assuming explicit preferences upfront, which is very close to the viewpoint of \say{learning} the preferences. For a dataset with finitely many observations, \say{generalized axiom of revealed preference} states the necessary and sufficient conditions are proposed for the existence of a well-behaved utility function that justifies the choices \cite{afriat1967construction}. When full rationalizability is not held, the \say{critical cost efficiency index} can quantify the degree of approximate rationalizability \cite{afriat1973system,polisson2024rationalizability}. The insistence on well-behavior and rationalizability for the inferred utility function is equivalent to assuming them as inductive biases for preference learning. While the former criterion is generally accepted, the latter assumes rationality on the part of the preference maker, which is usually the case in social and economic domains. 

Other efforts include practical attempts to obtain a preference index based on popular unsupervised data analysis methods, e.g., principal component analysis (PCA). One interesting example is developing indices of socio-economic status for an area, household, or individual. These measures are usually applied to analyze poverty, inequality, population classification, optimal public resource allocation, or to regress the index on parameters such as health status or economic behavior \cite{kolenikov2009socioeconomic}. The Australian Bureau of Statistics, suggests a socio-economic index for areas (SEIFA) which is widely used for academic researches and policy papers (see e.g. \cite{youens2025guide,clark2010western,wilkinson2001variation,beks2024application,people2015planning,bollen2007socio,mcintyre2000applying,goldie2014two,amir2008socioeconomic}). The mentioned index is constructed based on the first principal component of related basic social and economic variables \cite{ABS2021SEIFA} and the same technical details are presented for constructing the index of household advantage and disadvantage (IHAD) \cite{ABS2021IHAD}. Wealth indexes have quite similar use cases; they were proposed to capture wealth and poverty using proxy variables when straightforward numeric measures like income or consumption aren't available or reliable, which is common in developing economies \cite{kolenikov2009socioeconomic}. Filmer and Pritchett proposed a framework based on PCA \cite{filmer2001estimating}, which was later picked up by the World Bank \cite{gwatkin2007socio}. Comprehensive studies on quality of life try to go beyond GDP and capture a fuller picture of well-being, including health, education, living standards, employment, social exclusion and vulnerability, and access to communication and financial services. Corresponding composite indexes are usually referred to as multidimensional poverty (e.g., in \cite{alkire2015multidimensional,d2024multidimensional,tkach2022multidimensional}) or human development index \cite{human2019UNDP}.

The general idea of building a composite index by linear combination of basic indicators and using PCA to set the weights of the original variables is suggested as a simple, objective statistical approach \cite{maggino2017dealing,greco2019methodological}. From a technical point of view, the index construction schema mentioned, which is often adopted for multi-criteria assessments in social and economic domains, is similar to the warm-up stage of building a preference model with cold-start active learning. Although not common, the resulting preference model could be fine-tuned with expert supervision to achieve a better performance. The approach can be traced back to psychometric research seeking to compute intelligence quotient (IQ) \cite{terman1916measurement}, which summarizes positive correlations among cognitive tasks such as reasoning, planning, problem-solving, abstract thinking, comprehending complex ideas, and learning quickly from experience \cite{gottfredson1997mainstream}. The first principal component can be computed without any supervision and expresses an important, potentially interpretable latent variable \cite{kolenikov2009socioeconomic}. In addition, the projections on the principal component are robust to the presence of low-importance or redundant features. However, it has frequently been stated that, despite its practical performance, selecting the principal component as the desired research objective lacks a firm theoretical explanation, and the obtained weights should be interpreted with care \cite{bollen2002economic,bollen2007socio,human2019UNDP,ringner2008principal}.

\section{Framework} \label{sec:meth}

\begin{figure*}[!t]
  \centering
  \includegraphics[width=\textwidth]{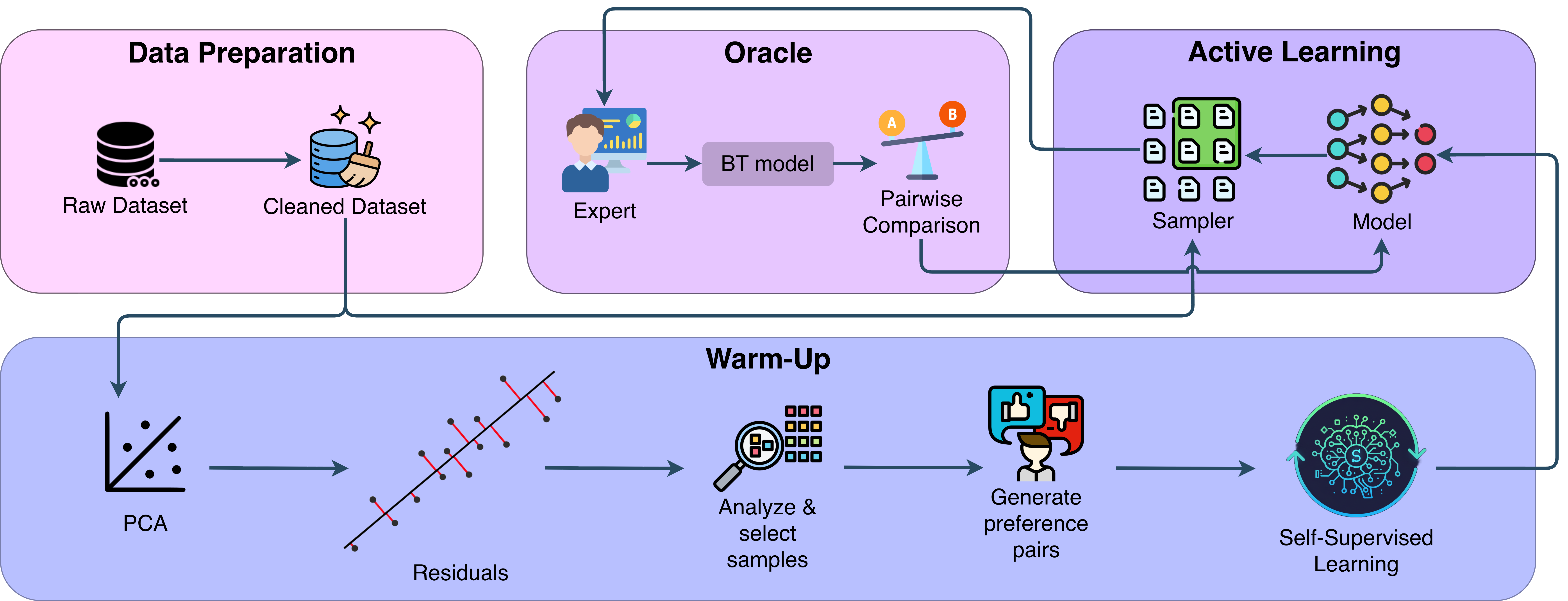}
  \caption{Conceptual overview of the cold-start active preference learning framework. All the elements and operations are described in detail within Section \ref{sec:meth}}
  \label{fig1}
\end{figure*}

This section details the proposed framework for cold-start active preference learning, designed to mitigate the issue in scenarios with no initial labeled data. Our framework, indicated in Figure \ref{fig1}, is structured into four distinct components in the following way: The initial phase focuses on processing the raw input dataset through basic cleaning operations, such as handling missing values and removing irrelevant features, to produce a refined and structured dataset suitable for subsequent analysis and model training. This next crucial phase bootstraps the learning process by first applying principal component analysis (PCA) to derive an unsupervised initial heuristic. Based on projection residuals along the principal component, a specific number of data pairs are stochastically sampled and labeled using their order along the principal component, forming a dataset for self-supervising and initializing a preference model. In this paper, we employ XGBoost \cite{chen2016xgboost}, a well-known gradient boosting implementation renowned for its state-of-the-art performance across a wide range of machine learning problems \cite{shmuel2024comprehensivebenchmarkmachinedeep}. Beginning with the warmed-up model, an iterative learning loop is implemented where a sampler selects informative data pairs to be labeled by the simulated oracle. The newly acquired preference labels are then used to train and refine the model, allowing it to improve its performance by leveraging the most beneficial data points. To emulate realistic expert feedback, the oracle simulation component describes the process of making comparative judgments based on the true underlying target values for each data point. These absolute values are processed by a Bradley-Terry (BT) model \cite{bradley1952rank}, which yields probabilistic preference results that mimic real-world labeling imperfections.

\subsection{Data Preparation}
Data preprocessing is a preliminary step to avoid unnecessary complications and ensure the data are suitable for subsequent analyses. The data cleaning and preprocessing stage involves several automatic (without supervision) and standard operations:
\begin{itemize}
    \item \textbf{Feature Selection:} Non-categorical string features are identified and removed.
    \item \textbf{Encoding Categorical Variables:} Based on a data-driven heuristic, categorical features were handled using one of two methods. Each categorical column was either converted into multiple binary features via one-hot encoding or mapped to a single numerical column.
    \item \textbf{Handling Missing Values:} Missing data (NaN values) are addressed, typically through imputation techniques or by removing instances/features with excessive missingness, depending on the dataset characteristics.
    \item \textbf{Data Normalization/Standardization:} Feature values are standardized to have zero mean and unit variance to prevent features with larger magnitudes from unduly influencing model training.
\end{itemize}
Upon completion of these operations, we obtain a cleaned dataset, denoted as $\mathcal{D} = \{(\mathbf{x}_i, y_i)\}_{i=1}^n$. Each instance $i$ is represented by a feature vector $\mathbf{x}_i \in \mathbb{R}^p$, where $p$ is the dimensionality after feature selection, and $y_i \in \mathbb{R}$ is an associated scalar value (e.g., a utility score, a performance metric, or an objective function value) from which preferences can be derived.

\subsection{Warm-Up} \label{sec:warm}
This phase bootstraps the preference learning process in the absence of initial expert labels. It involves leveraging the inherent structure of the cleaned data through Principal Component Analysis (PCA) \cite{hastie2009} to generate surrogate preference labels, sampling data based on data errors, and then pre-training an initial model in a self-supervised manner. The resulting \say{warmed-up} model, $\mathcal{M}_0$, provides an informed starting point for subsequent active learning.

\subsubsection{PCA for Trend Approximation and Surrogate Score Generation}
Given the cleaned feature matrix $\mathbf{X} \in \mathbb{R}^{n \times p}$ (where rows represent data points and columns represent features, assumed to be column-centered), we apply PCA to identify the primary axis of variation. This approach is motivated by the established use of the first principal component in constructing composite socio-economic indexes, where it serves as a proxy for an underlying latent construct by capturing the maximum variance across a set of observable indicators (see, e.g., \cite{filmer2001estimating,vyas2006}). The first principal component, represented by the weight vector $\mathbf{w} \in \mathbb{R}^p$, is found by solving the optimization problem:
\[
\mathbf{w} = \arg\max_{\|\mathbf{w}\|_2=1} \mathbf{w}^\top \mathbf{S} \mathbf{w} = \arg\max_{\|\mathbf{w}\|_2=1} \frac{1}{n} \mathbf{w}^\top \mathbf{X}^\top \mathbf{X} \mathbf{w},
\]
where $\mathbf{S} = \frac{1}{n}\mathbf{X}^\top \mathbf{X}$ is the sample covariance matrix. The vector $\mathbf{w}$ is the principal eigenvector of $\mathbf{S}$ corresponding to its largest eigenvalue. The projection of each data point $\mathbf{x}_i$ (representing the $i$-th row of $\mathbf{X}$) onto this principal component yields a scalar score $t_i$:
\[
\mathbf{t} = \mathbf{X} \mathbf{w}, \quad \text{thus } t_i = \mathbf{x}_i \mathbf{w}.
\]
The vector $\mathbf{t} \in \mathbb{R}^n$ contains these one-dimensional projections, which serve as surrogate values for generating initial preference labels.

The rationale for employing a single principal component as a surrogate for underlying labels, particularly in socio-economic contexts, is grounded in several observations:
\begin{itemize}
    \item \textbf{Popularity and Practical Success:} As noted in section \ref{sec:prev}, PCA-derived composite indexes are conventional in social and economic domains. They leverage simplicity, interpretability, objectivity, and reasonable judgment.
    \item \textbf{Revelation of Homothetic Preferences:} A preference model is called homothetic if it can be represented by a utility function which is homogeneous of degree 1 \cite{varian1992microeconomic}. Under this setting, the maximal attainable utility (i.e., the indirect utility function) can be written as a linear function of wealth. Assuming rationality - which is admissible in many social and economic datasets - PCA would be capable of retrieving the revealed indirect utility function. Therefore, employing the first principal component can be interpreted as approximating the target preference model with the aligned internal preferences of dataset entities, which constitutes a considerable inductive bias.
    \item \textbf{Maximization of Variance:} PCA identifies the direction in the feature space that captures the maximum variance. If an underlying latent factor (e.g., overall quality, development level, or a composite socio-economic indicator) significantly influences multiple observed features, it is likely to align with the direction of the highest variance. The first principal component score $t_i$ can thus be interpreted as an estimate of this dominant latent variable for each data point $i$.
    \item \textbf{Linear Aggregation Proxy:} Many socio-economic indexes or utility functions are constructed as linear combinations (often weighted averages) of various indicators. The PCA score $t_i = \sum_{j=1}^p x_{ij} w_j$ is itself a linear combination of the features. If the true underlying labels $y_i$ are, or can be approximated by, such a linear aggregation, the first PC provides a data-driven method to estimate these weights $w_j$ and the resulting aggregate scores $t_i$ in an unsupervised manner.
    \item \textbf{Objective Initial Ordering Criteria:} In the absence of expert labels (the cold start scenario), PCA offers an objective data-driven approach to establish an initial ordering based on the most prominent pattern within the dataset. This can be more robust than random initialization or heuristics that do not consider the data distribution.
\end{itemize}

\subsubsection{Tuning the Strength of the Prior}
To assess how well our assumption—that preferences are revealed by a linear indirect utility function—holds, we analyze the principal component reconstruction error. The data points $\mathbf{X}$ can be approximated by projecting them onto the line defined by $\mathbf{w}$ and then mapping back to the original $p$-dimensional space:
\[
\hat{\mathbf{X}} = \mathbf{t}\,\mathbf{w}^\top.
\]
The residual for each data point $\mathbf{X}_i$ is the Euclidean distance between the original point and its reconstruction:
\[
r_i = \|\mathbf{X}_i - \hat{\mathbf{X}}_i\|_2 = \|\mathbf{X}_i - (\mathbf{X}_i \mathbf{w})\mathbf{w}^\top\|_2.
\]
The average squared residual, $\sigma_r^2 = \widehat{\mathrm{Var}}(r) = \frac{1}{n} \sum_{i=1}^n r_i^2$, quantifies the variance not captured by the first principal component and serves as an indicator of the overall PCA fit.

The number of pairwise samples used for pre-training, $N_{\text{pre}}$, is determined dynamically based on $n$, $\sigma_r^2$, and hyperparameters $k$ and $\alpha$:
\[
N_{\text{pre}} = \left\lfloor \frac{n \cdot k}{1 + \alpha \cdot \sigma_r^2} \right\rfloor.
\]
Here, $k \in [1, 100]$ scales $N_{\text{pre}}$ relative to $n$, while $\alpha \in [10^{-7}, 10^{-4}]$ modulates the influence of $\sigma_r^2$. A lower $\sigma_r^2$ (better PCA fit) leads to a larger $N_{\text{pre}}$, encouraging more reliance on PCA-derived labels.

\subsubsection{Residual-Based Pair Generation}
The $N_{\text{pre}}$ pairs for model initialization are constructed using a weighted sampling strategy that prioritizes data points well-represented by PCA. The selection probability $p_s(\mathbf{x}_i)$ for an individual data point $\mathbf{x}_i \in \mathcal{D}$ is inversely proportional to its residual $r_i$:
\[
p_s(\mathbf{x}_i) \propto \frac{1}{r_i + \epsilon}, \quad \text{such that} \sum_{i=1}^n p_s(\mathbf{x}_i) = 1,
\]
where $\epsilon > 0$ is a small constant for numerical stability. Pairs $(\mathbf{x}_i, \mathbf{x}_j)$ are formed by drawing two distinct data points from $\mathcal{D}$ according to these probabilities.

Using the PCA-derived scores $t_i$ and $t_j$ for the sampled pair, a pairwise pseudo-label $\ell^{\text{PCA}}_{ij}$ is generated:
\[
\ell^{\text{PCA}}_{ij} =
\begin{cases}
1, & \text{if } t_i > t_j \quad (\text{indicating } \mathbf{x}_i \succ_{\text{PCA}} \mathbf{x}_j), \\
0, & \text{if } t_i \le t_j \quad (\text{indicating } \mathbf{x}_j \succeq_{\text{PCA}} \mathbf{x}_i).
\end{cases}
\]
This process creates the \textit{PCA-labeled preference dataset}, $\mathcal{D}_P^{\text{PCA}} = \{([\mathbf{x}_u, \mathbf{x}_v], \ell^{\text{PCA}}_{uv})\}_{m=1}^{N_{\text{pre}}}$.

\subsubsection{Self-Supervised Model Initialization}
The preference learning model, an XGBoost binary classifier (\texttt{XGBClassifier}) in our framework, is then pre-trained using this PCA-labeled dataset $\mathcal{D}_P^{\text{PCA}}$. This stage is considered self-supervised as no human-annotated labels are involved; the labels are derived solely from the data's internal structure via PCA. The model is trained by minimizing the logistic loss (binary cross-entropy):
\begin{multline} \label{eq:logloss_coldstart}
\mathcal{L}(\boldsymbol\theta) = - \sum_{m=1}^{N_{\text{pre}}}
\Bigl[
\ell^{\text{PCA}}_{uv_m} \ln p_{\boldsymbol\theta}(\mathbf{z}_{uv_m})
\\
+
(1 - \ell^{\text{PCA}}_{uv_m}) \ln(1 - p_{\boldsymbol\theta}(\mathbf{z}_{uv_m}))
\Bigr],
\end{multline}
where $\mathbf{z}_{uv_m} = [\mathbf{x}_u; \mathbf{x}_v]_m$ is the concatenated feature vector of the $m$-th pair, and $p_{\boldsymbol\theta}(\mathbf{z}_{uv_m})$ is the probability predicted by the XGBoost model with parameters $\boldsymbol\theta$ that $\mathbf{x}_u$ is preferred to $\mathbf{x}_v$.
This pre-training yields the warmed-up model $\mathcal{M}_0$, which is equipped with an initial understanding of preferences before interacting with the (simulated) expert oracle.

\subsection{Warm-Start Active Learning}
The final phase of our framework implements an active learning loop to iteratively refine the preference learning model. This process commences with the warmed-up model, $\mathcal{M}_0$, which is then progressively updated by strategically acquiring and incorporating new preference labels generated by the simulated expert oracle.

The iterative active learning process at each step $t$ (for $t=1, 2, \dots, T_{\text{max}}$) proceeds as follows:

Let $\mathcal{M}_{t-1}$ be the state of the XGBoost preference learning model at the beginning of iteration $t$.
\begin{enumerate}
    \item \textbf{Training Batch Request and Query Strategy Formulation:}
    The model $\mathcal{M}_{t-1}$ signals to the \say{sampler} component that it is ready for a new training batch. This involves the model providing information about its current state or uncertainty levels. The sampler is then tasked with selecting a batch of $N_b$ unlabeled pairs $(\mathbf{x}_u, \mathbf{x}_v)$ from the overall pool of available unlabeled pairs. The selection strategy employed by the sampler can be configured to use either random sampling or an uncertainty-based sampling approach, in which pairs for which the current model $\mathcal{M}_{t-1}$ exhibits high uncertainty are prioritized.

    \item \textbf{Oracle Label Acquisition via Sampler-Expert Interaction:}
    The sampler communicates the selected batch of $N_b$ pairs to the expert component within the simulated oracle system (detailed in Section \ref{sec:oracle}). For each queried pair $(\mathbf{x}_u, \mathbf{x}_v)$, the oracle executes its internal process: the expert ascertains the true underlying target values for $\mathbf{x}_u$ and $\mathbf{x}_v$, and then the Bradley-Terry model generates a (potentially noisy) preference label $\ell_{uv}^{\text{oracle}}$. This results in a newly oracle-labeled batch $\mathcal{B}_t = \{([\mathbf{x}_u, \mathbf{x}_v], \ell_{uv}^{\text{oracle}})\}_{k=1}^{N_b}$.

    \item \textbf{Incremental Model Update:}
    The XGBoost model $\mathcal{M}_{t-1}$ is then updated using the newly acquired oracle-labeled batch $\mathcal{B}_t$. XGBoost's incremental training capability is leveraged, allowing the model to build on its existing state. If $\boldsymbol\theta_{t-1}$ represents the parameters of the model $\mathcal{M}_{t-1}$ (or its internal tree ensemble), the update to obtain $\mathcal{M}_t$ can be conceptualized as:
    \[
    \begin{aligned}
      \mathcal{M}_t = \text{XGBoost.train}(&\text{objective}=\text{logistic}, \\
                                           &\text{params}=\boldsymbol{\theta}_{t-1}, \\
                                           &\text{training\_data}=\mathcal{B}_t, \\
                                           &\text{existing\_model}=\mathcal{M}_{t-1})
    \end{aligned}
    \]
    More specifically, when utilizing XGBoost's learning API, one can continue training an existing model object by providing it with new data batches, effectively refining the model parameters $\boldsymbol\theta$ to $\boldsymbol\theta_t$.
\end{enumerate}
This iterative cycle of the model requesting data via the sampler, the sampler querying the oracle, the oracle providing labels, and the model subsequently updating, continues for a predefined number of iterations $T_{\text{max}}$, or until a specific budget for oracle queries is exhausted. This incremental and adaptive approach enables the model to efficiently integrate expert feedback and enhance its preference-prediction capabilities over time.

\subsection{Oracle Simulation} \label{sec:oracle}
To emulate the process of obtaining preference labels from a human expert—a process often characterized by inherent inconsistencies or stochasticity—we implement a simulated expert oracle. This oracle generates realistic, potentially noisy preference labels for pairs of data points. The process begins when two data points, say $\mathbf{x}_i$ and $\mathbf{x}_j$, are fetched from the cleaned dataset via a component named \say{sampler}, which will further be explained. An \say{expert} component within the oracle then ascertains the true underlying target values, $y_i$ and $y_j$, for these respective data points. These true target values are subsequently used by a Bradley-Terry (BT) model to probabilistically generate a preference label $\ell_{ij}^{\text{oracle}}$ for the pair $(\mathbf{x}_i, \mathbf{x}_j)$, thereby incorporating realistic noise into the labeling process. These generated preference labels are then provided to the model for its training updates.

\subsubsection{Probabilistic Preference Generation via Bradley-Terry Model}
The core of our simulated oracle's noise model is the Bradley-Terry (BT) model \cite{bradley1952rank}. The BT model is a well-established statistical framework specifically designed for analyzing pairwise comparison data. It posits that the probability of one item $i$ being preferred over another item $j$ (denoted $\mathbf{x}_i \succ \mathbf{x}_j$) can be expressed in terms of underlying positive-valued \say{strength} parameters, $\beta_i$ and $\beta_j$, associated with each item.

In our simulation, once the expert component has provided the true target values $y_i$ and $y_j$ for the items $\mathbf{x}_i$ and $\mathbf{x}_j$, their respective strength parameters $\beta_i$ and $\beta_j$ are derived. Typically, $\beta_k$ is directly proportional to $y_k$, i.e., $\beta_k = f(y_k)$, where $f$ is a function ensuring $\beta_k > 0$. For example, if all $y_k$ values are already positive, $f(y_k)$ can simply be $y_k$ itself, which is the case for this work. If $y_k$ values might be negative or zero, a suitable transformation (such as scaling and shifting, or the exponentiation discussed subsequently) must be applied first. The probability that the oracle then deems item $\mathbf{x}_i$ preferable to item $\mathbf{x}_j$ is formally given by the following expression:
\[
\Pr(\mathbf{x}_i \succ \mathbf{x}_j | y_i, y_j) = \frac{\beta_i}{\beta_i + \beta_j}.
\]
This probabilistic formulation ensures that even when starting from known true underlying values $y_i$ and $y_j$, the final preference label is generated stochastically. This approach realistically reflects the inherent fallibility or variability often observed in expert judgments.

\subsubsection{Handling Exponentially Scaled Target Values}
In scenarios where the true underlying target values $y_i$ (provided by the expert component) are known or assumed to follow an exponential scale, or perhaps where their differences are more meaningfully interpreted on a logarithmic scale (e.g., certain types of utility scores where ratios matter), a common variant of the BT model is employed. This variant is specifically tailored for generating the preference probability under such conditions. In this approach, the strength parameters are defined using the exponential of the scores, $\beta_k = e^{s_k}$, where $s_k$ represents the score of item $k$ (often $s_k = y_k$ or potentially a linear transformation of $y_k$). The resulting probability of preference calculation then becomes:
\[
\Pr_{\text{exp}}(\mathbf{x}_i \succ \mathbf{x}_j | y_i, y_j) = \frac{e^{s_i}}{e^{s_i} + e^{s_j}} = \frac{1}{1 + e^{-(s_i - s_j)}}.
\]
This expression is mathematically equivalent to applying a standard logistic function (sigmoid) to the difference in scores $s_i - s_j$. The decision regarding whether to use the standard BT formulation (where $\beta_k \propto y_k$) or this alternative exponential variant is primarily guided by the assumed nature and scale of the target values $y_i$, as understood for the specific dataset or application domain under consideration.

Regardless of the specific formulation chosen (standard or exponential), the oracle ultimately generates the final binary preference label $\ell_{ij}^{\text{oracle}}$ for the given pair $(\mathbf{x}_i, \mathbf{x}_j)$. This is achieved by sampling from a Bernoulli distribution, which is parameterized by the calculated preference probability, either $\Pr(\mathbf{x}_i \succ \mathbf{x}_j | y_i, y_j)$ or $\Pr_{\text{exp}}(\mathbf{x}_i \succ \mathbf{x}_j | y_i, y_j)$.

\begin{figure}[H]
    \centering
    \includegraphics[width=\columnwidth]{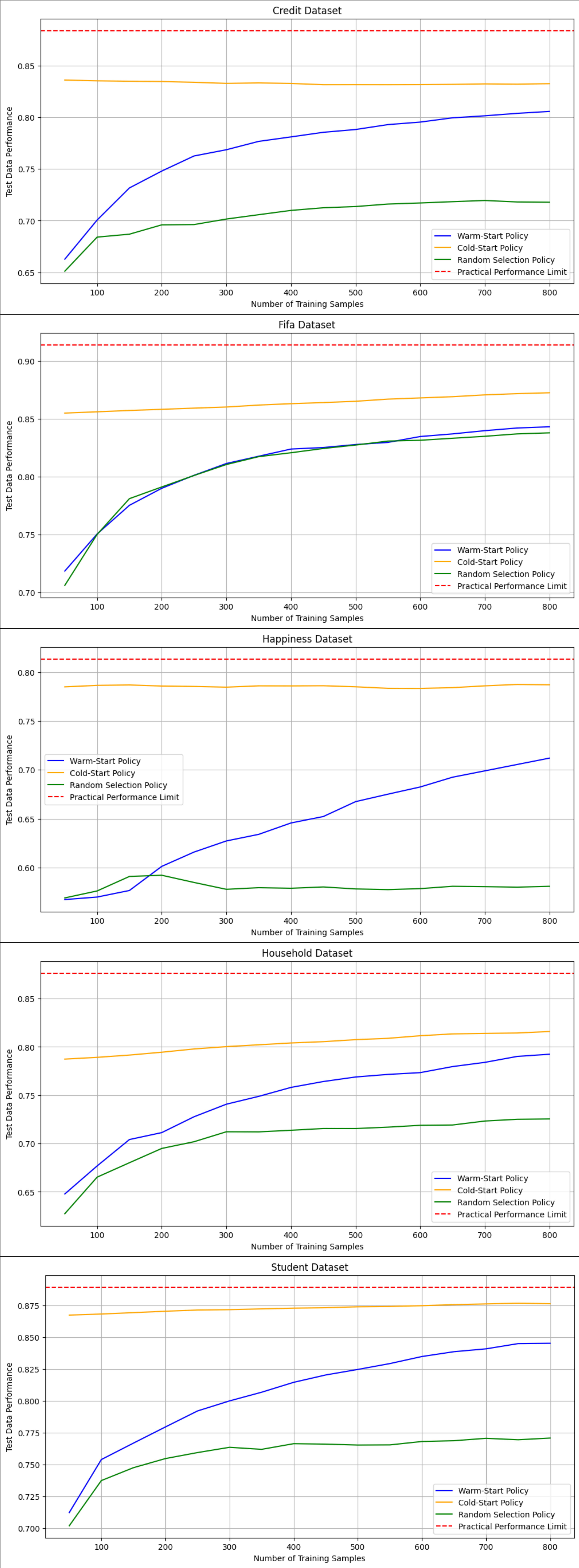}
    \caption{Comparative performance across different datasets (Averaged over 40 runs).}
    \label{fig:low_data_results}
\end{figure}

\section{Experiments}
This section presents the empirical evaluation of the proposed cold-start active preference learning framework. The experimental setup, datasets, compared learning approaches, and evaluation protocol used to assess the framework's efficacy are described.

\subsection{Experimental Setup} To ensure a fair and consistent evaluation, the experimental setup was standardized across all datasets. We employed an XGBoost binary classifier as our preference learning model. For evaluation, a common test set was constructed for each dataset by randomly sampling 20,000 unique pairs of instances, and all reported accuracy scores are calculated on this set. Key model hyperparameters were configured as follows: the number of boosting rounds was set to 500, and the maximum tree depth was determined dynamically for each dataset based on its feature dimensionality via the formula: \[ \text{tree\_depth} = \text{round}\left(\sqrt{N_{\text{features}}}\right). \] where $N_{\text{features}}$ is the number of columns in the processed data. All other XGBoost parameters were kept at their default values to isolate the impact of the different learning policies and the cold-start initialization.

\subsection{Datasets}
To demonstrate the generalizability and robustness of our framework, experiments were conducted on different datasets, each representing a distinct domain and varying characteristics:
\begin{itemize}
    \item \textbf{Credit Socring:} This dataset contains information related to the credit scores of bank clients, with the objective of learning preferences over client profiles.
    \item \textbf{Football Player Market Price:} Comprising data on football players from the Fifa 2022 dataset, where the target value for preference elicitation is the market price of the players.
    \item \textbf{Happiness Index 2019:} This dataset includes national happiness rankings from 2019, based on multiple contributing factors, used here to learn preferences between countries.
    \item \textbf{Household Welfare:} A large-scale dataset (more than 1.5 million data points) detailing various socio-economic variables pertinent to the welfare of Iranian households.
    \item \textbf{Student Performance:} Includes academic performance data (grades) of students, correlated with factors like family background, study time, and ethnicity.
\end{itemize}

\subsection{Comparison with an Alternative Initialization Method}
To further validate the effectiveness of our simple yet powerful PCA-based initialization, we compare it with a strong unsupervised benchmark derived from modern large language models and matrix factorization techniques. This method, which we term the \say{GPT-based initialization}, involves a sophisticated pipeline where features are first preprocessed, and then a consensus utility score is generated from the outputs of both Nonnegative Matrix Factorization (NMF) and Singular Value Decomposition (SVD). This final utility score is then used to create a pre-training dataset in the same manner as our PCA-based score.

We evaluated the quality of both initialization methods by pre-training a separate model using each approach and then testing its performance directly on the common test set, without any subsequent active learning. Table \ref{tab:init_comparison} presents the resulting pairwise accuracies for five datasets.

\begin{table}[h!]
\centering
\caption{Pairwise accuracy (\%) on the test set immediately after pre-training, comparing our proposed PCA-based initialization against the GPT-based alternative.}
\label{tab:init_comparison}
\begin{tabular}{lcc}
\hline
\textbf{Dataset} & \textbf{PCA Pre-trained (\%)} & \textbf{GPT Pre-trained (\%)} \\ \hline
Credit & \textbf{84.82} & 79.80 \\
Fifa & 85.45 & \textbf{87.67} \\
Happiness & \textbf{78.74} & 76.00 \\
Household & \textbf{79.09} & 76.53 \\
Student & 87.09 & \textbf{88.25} \\ \hline
\end{tabular}
\end{table}

The results in Table \ref{tab:init_comparison} show that our proposed PCA-based initialization is highly competitive and generally outperforms the more complex GPT-based benchmark. On three of the five datasets (Credit, Happiness, and Household), our simpler method provides a better warm start. In the two cases where the GPT-based method performed better (Fifa and Student), the margin of improvement was minimal. This comparison validates that our computationally efficient PCA approach is a robust and effective strategy for initializing the model, achieving performance comparable to or superior to a strong alternative benchmark.

\subsection{Compared Learning Policies and Benchmark}
To evaluate the effectiveness of our proposed framework, we compare three distinct learning policies and establish a benchmark named \say{Practical Performance Limit} to contextualize the results:

\begin{enumerate}
    \item \textbf{Random Selection Policy:} This policy serves as a fundamental baseline. The XGBoost model is initialized from scratch (a blank state), and in the active learning phase, pairs are selected at random from the available pool for oracle labeling. This approach is depicted in green in our visual results.

    \item \textbf{Warm-start Policy:} This policy represents a classic active learning approach. It misses any treatment for pre-training and cold-start issues. The model is initialized from scratch and employs an uncertainty-based sampling strategy to select pairs for oracle labeling. This approach is depicted in blue.

    \item \textbf{Cold-start Policy:} This is our proposed full framework, designed specifically to address the cold start problem where no initial labels exist. The XGBoost model is first initialized using the unsupervised warm-up procedure detailed in Section \ref{sec:warm}. Subsequently, in the active learning phase, it uses uncertainty sampling to select the most informative pairs for oracle labeling. This approach is depicted in orange.
    
    \item \textbf{Practical Performance Limit:} To establish a \say{Practical Performance Limit}, we created a near-saturated benchmark model for each dataset. This was achieved by iteratively training a model on a large budget of 100,000 pairs, starting with an initial random batch of 1000 pairs and then using uncertainty sampling for the 99 subsequent batches. The resulting accuracy provides a strong empirical upper-bound for comparison and is depicted in red in our plots.
\end{enumerate}
All three policies utilize an incremental training paradigm, where the model is continuously updated with new batches of oracle-labeled data.

\subsection{Evaluation Scenario and Metric}
The primary evaluation metric used across all experiments is **Pairwise Accuracy**, calculated on the aforementioned common test set. This metric directly measures the proportion of correctly predicted preference pairs ($i \succ j$), providing a clear and relevant assessment for the preference learning task. We investigate the performance of the compared policies in an experimental scenario designed to rigorously evaluate learning efficiency under scarce labeled data.

This **Low-Data Regime** scenario focuses on the critical early stages of learning. The models are incrementally trained on oracle-labeled data, starting with 50 pairs and increasing in steps of 50 pairs to a total of 800 pairs. To ensure robust conclusions and mitigate the effects of random sampling variability, this entire process is repeated 40 times for each policy on each dataset. The performance curves reported in our results represent the average pairwise accuracy across these 40 independent runs.

The results of this evaluation are presented in Figure \ref{fig:low_data_results}. The figure illustrates the performance of the three compared learning policies, plotting the average pairwise accuracy as a function of the number of oracle queries. The plots clearly demonstrate the superior sample efficiency and the immediate impact of our proposed cold-start strategy. Across all evaluated datasets, our **Cold-start Policy** consistently establishes a significant performance advantage over the baseline policies, achieving higher accuracy with substantially fewer labeled pairs.

\section*{Data and Code Availability Statement}

The source code for this study, including the implementation of the proposed framework and scripts for replicating the experiments, is openly available. The repository is hosted on GitHub at: \url{https://github.com/Dan-A2/cold-start-preference-learning}.

All datasets utilized in this research are publicly available.

\section{Conclusion}
This paper addresses a critical challenge in the practical application of preference learning: the cold-start problem in active learning. This issue, characterized by poor performance in the absence of an initial labeled dataset, has limited the adoption of active learning in real-world scenarios, especially in social and economic research, where obtaining labeled data is a significant bottleneck.

To address this limitation, the proposed framework introduces a self-supervised warm-up phase that employs Principal Component Analysis (PCA) to generate initial pseudo-labels, thereby providing a foundational model without the need for hand-labeled data. This model is then incrementally refined through an active learning loop that strategically queries a simulated noisy oracle, replicating a realistic human-in-the-loop process.

Empirical evaluation across diverse datasets confirms the efficacy of the proposed method. The cold-start policy consistently and significantly outperforms standard active learning strategies, achieving higher accuracies with substantially fewer labeled data in low-data scenarios. This improved sample efficiency enhances the practicality and cost-effectiveness of active preference learning for researchers in social and economic domains, reducing barriers to the application of advanced preference modeling in complex societal contexts.

Future research could extend this work in several directions. Although PCA has proven effective for initialization, exploring alternative unsupervised or self-supervised techniques may yield further improvements. Additionally, developing more advanced active-learning pair-selection strategies that leverage information obtained during the warm-up phase represents a promising avenue. Finally, applying and validating the framework in real-world case studies involving human experts would provide critical evidence of its practical benefits beyond simulated environments.

\bibliographystyle{unsrt}
\bibliography{main}

\end{document}